\documentclass[pdflatex,sn-mathphys-num]{sn-jnl}

\usepackage{graphicx}%
\usepackage{multirow}%
\usepackage{amsmath,amssymb,amsfonts}%
\usepackage{amsthm}%
\usepackage{mathrsfs}%
\usepackage[title]{appendix}%
\usepackage{xcolor}%
\usepackage{textcomp}%
\usepackage{manyfoot}%
\usepackage{booktabs}%
\usepackage{algorithm}%
\usepackage{algorithmicx}%
\usepackage{algpseudocode}%
\usepackage{listings}%

\theoremstyle{thmstyleone}%
\theoremstyle{thmstyletwo}%

\theoremstyle{thmstylethree}%

\begin{document}

\title[MML Nature]{Method of Manufactured Learning for Solver-free Training of Neural Operators}


\author[1]{\fnm{Arth} \sur{Sojitra}} \email{asojitra@vols.utk.edu}
\author*[1]{\fnm{Omer} \sur{San}}\email{osan@utk.edu}

\affil*[1]{\orgdiv{Department of Mechanical and Aerospace Engineering}, \orgname{University of Tennessee}, \orgaddress{\city{Knoxville}, \state{TN}, \country{USA}}}

\abstract{Training neural operators to approximate mappings between infinite-dimensional function spaces often requires extensive datasets generated by either demanding experimental setups or computationally expensive numerical solvers. This dependence on solver-based data limits scalability and constrains exploration across physical systems. Here we introduce the Method of Manufactured Learning (MML), a solver-independent framework for training neural operators using analytically constructed, physics-consistent datasets. 
Inspired by the classical method of manufactured solutions, MML replaces numerical data generation with functional synthesis, i.e., smooth candidate solutions are sampled from controlled analytical spaces, and the corresponding forcing fields are derived by direct application of the governing differential operators. During inference, setting these forcing terms to zero restores the original governing equations, allowing the trained neural operator to emulate the true solution operator of the system. The framework is agnostic to network architecture and can be integrated with any operator learning paradigm. In this paper, we employ Fourier neural operator as a representative example. Across canonical benchmarks including heat, advection, Burgers, and diffusion-reaction equations. MML achieves high spectral accuracy, low residual errors, and strong generalization to unseen conditions. By reframing data generation as a process of analytical synthesis, MML offers a scalable, solver-agnostic pathway toward constructing physically grounded neural operators that retain fidelity to governing laws without reliance on expensive numerical simulations or costly experimental data for training.}


\maketitle

\section{Introduction}

Advances in artificial intelligence are reshaping the scientific enterprise, promising systems that can uncover structure in complex data and accelerate discovery across physics, biology, and engineering. Yet the rapid scaling that drives modern AI, where increasingly large models are fueled by massive and easily collected datasets, stands in clear tension with the realities of scientific applications. High-fidelity experimental measurements and large ensemble numerical simulations are often expensive, limited, or entirely inaccessible \cite{ghattas2021learning}. This mismatch has intensified the need for learning strategies that do not depend on vast supervised datasets but instead draw power from the mathematical structure of physical laws. Within this landscape, self-supervised and physics-aware learning frameworks are becoming essential for building scientific AI systems that can generalize beyond scarce data, remain faithful to governing equations, and operate reliably in regimes where traditional supervision cannot be obtained. 

A central motivation for data-driven modeling in the sciences is the construction of surrogate systems that can replace or augment classical high-fidelity solvers. In practice, building such surrogates, reduced-order models, and latent-space emulators requires a substantial conversion process: large ensembles of numerical solutions must first be generated by legacy solvers, and these solver outputs are then distilled into a trainable representation that enables rapid inference. The offline phase is often computationally intense, involving thousands of simulations or even far more to adequately sample the relevant parametric or functional spaces. Although the resulting models can deliver orders of magnitude speed-ups during inference, the heavy cost of dataset generation is rarely emphasized, and in many cases it becomes the dominant bottleneck in the development pipeline \cite{ahmed2021closures}. This challenge becomes even more critical in the design of digital twins for scientific applications \cite{Willcox2023,ferrari2024digital,Stadtmann2023}, where near-real-time predictive capability is not only desirable but essential. Together, these considerations underscore a fundamental need in scientific machine learning: to preserve the efficiency and flexibility of learned surrogates while reducing or eliminating reliance on expensive solver-produced data.

In response, learning solution operators for partial differential equations (PDEs) has emerged as a compelling alternative to traditional numerical solvers, enabling mesh-independent generalization and fast inference capability across families of physical systems.
Operator Learning~\cite{kovachki2023neural, azizzadenesheli2024neural, kovachki2024operator, subedi2025operator} provides a framework for approximating mappings between infinite-dimensional function spaces. Early foundational architectures such as the Deep Operator Network (DeepONet)~\cite{lu2021learning} and the Fourier Neural Operator (FNO)~\cite{li2020fourier1} have demonstrated the feasibility of learning such mappings directly from data. Building on this foundation, a diverse ecosystem of neural operators has emerged, encompassing graph-based formulations that capture geometric dependencies~\cite{li2020multipole, sharma2024graph, li2020neural, cho2024graphdeeponet}, convolutional and attention based architectures~\cite{raonic2023convolutional, kissas2022learning, calvello2024continuum, hao2023gnot, guibas2021adaptive, li2022transformer, bryutkin2024hamlet}, wavelet and spectral-localized operators~\cite{tripura2023wavelet, lei2025u}, multiscale and hierarchical  frameworks~\cite{liu2022ht, luo2024hierarchical, jiang2025hierarchical}, resolution-independent frameworks \cite{bahmani2025resolution}, derivative-informed operators \cite{o2024derivative, luo2025efficient, cao2025derivative}, and probabilistic neural operators \cite{kutyniok2025probabilistic} among others.

Despite these advances, the effectiveness of operator-learning frameworks remains fundamentally constrained by the availability and quality of training data~\cite{viswanath2023neural, chen2024data, de2023convergence, lu2022comprehensive, kossaifimulti, li2025generalizability, zhou2406strategies}. In many physical systems, these datasets also contain strong correlations, so increasing their size does not necessarily yield a proportional gain in informational content \cite{ahmed2023dual}. Most current approaches rely on datasets generated through conventional numerical solvers that are computationally intensive and often restricted to narrow parametric regimes. As a result, the learned operators frequently inherit solver-induced biases, including numerical dissipation, dispersion, aliasing, and discretization or resolution-dependent errors all of which can hinder their ability to generalize to unseen physical regimes~\cite{chatain2025numerical, lippe2308pde,zhang2023label, liu2023domain, pfaff2020learning}. Furthermore, the reliance on solver-derived data imposes significant scalability limitations, especially for nonlinear, multiscale, or stiff systems that demand prohibitively fine discretizations to remain stable~\cite{kontolati2024learning,li2024m2no,george2022incremental}. To address these challenges, recent research has advanced physics-aware and hybrid operator-learning paradigms that embed governing equations within the training objective through weak-form and variational formulations~\cite{xu2024variational,patel2024variationally} and physics-informed regularization~\cite{raissi2019pinn, karniadakis2021physics, goswami2023physics, wang2021learning, li2024physics, toscano2025pinns}. Complementary developments including orthogonal polynomial expansions, boundary-aware kernels, and constraint-preserving formulations further mitigate the boundary inconsistencies by embedding boundary conditions directly into the operator architecture, thereby extending applicability to non-periodic and complex geometries~\cite{liu2024render, saad2022guiding, zhang2023label, boya2024physics}.

These challenges have motivated an emerging class of synthetic, solver-free data generation strategies, where one constructs solution-forcing pairs directly from symbolic or differentiable representations of the governing equations. A notable direction is the “backward” synthesis paradigm~\cite{roache2002code}, in which smooth candidate solutions \(u\) are sampled from controlled functional spaces and the corresponding forcing terms 
\(f\) are obtained analytically by applying the differential operator. Hasani and Ward~\cite{hasani2025synthetic} demonstrated this approach for elliptic equations using eigenfunction expansions. Related ideas have appeared in differentiable and symbolic physics frameworks~\cite{rackauckas2020universal1,cranmer2020lagrangian1,lusch2018deep1,greydanus2019hamiltonian1}, and operator-theoretic synthesis of function–response pairs for model discovery~\cite{rudy2017data1,champion2019data1,brunton2022data1}.  These efforts offer a promising alternative to solver-dependent datasets by providing mathematically controlled, noise-free data.

However, existing backward-manufactured frameworks are inherently spatial and static and lack explicit temporal dimension and thus cannot capture the causal and dissipative evolution intrinsic to parabolic or hyperbolic systems. Such formulations do not explicitly enforce energy conservation, entropy growth, or dissipation laws across time~\cite{eyink2006onsager1,boulle2023elliptic, diab2025temporal}. Moreover, without residual-consistent constraints during training, neural operators fitted purely to synthetically paired spatial data can exhibit physically inconsistent transients such as violating monotonic energy decay in diffusion~\cite{hu2024energetic1,zhang2024energy11} or entropy production in viscous Burgers and Navier-Stokes flows~\cite{van2024energy}. Several recent efforts have attempted to incorporate temporal structure and physical invariants into operator learning. Neural differential equation formulations~\cite{chen2018neural1,dupont2019augmented1,jia2019neural,quaglino2019snode} offer time-continuous representations and enable end-to-end differentiability across trajectories, yet they primarily capture dynamics at the state level and seldom enforce consistency with the governing partial differential equations. 

This leaves a critical gap between symbolic manufacturing, solver-free data generation, and time-dependent operator learning. The Method of Manufactured Solutions (MMS) offers conceptual grounding for bridging this divide. MMS was originally developed for code verification and has since been widely employed to validate legacy PDE solvers across a broad range of scientific and engineering domains~\cite{knupp2002verification, salari2000code, oberkampf2002verification}. One prescribes an analytical solution to a PDE and computes the associated forcing term so that the manufactured solution satisfies the equation exactly. This procedure enables rigorous assessment of numerical correctness, free from modeling uncertainties. In the present paper, we extend this philosophy to scientific machine learning by introducing the Method of Manufactured Learning (MML), a solver-independent framework that constructs training datasets by prescribing analytical solution families and deriving corresponding forcing fields through exact application of the governing operator. Because the solutions and their derivatives are known analytically, MML provides explicit control over smoothness, spectral richness, and functional complexity, enabling precise characterization of operator-learning performance. In doing so, MML transforms the MMS paradigm from a tool for verifying numerical solvers into a general methodology for training and evaluating neural operators.

By generating analytical, temporally consistent solution-forcing pairs, MML embeds the structure of the governing PDE directly into the training process. This reframes data generation as a problem of functional synthesis rather than numerical simulation, providing a scalable, physically grounded alternative to solver-derived datasets and enabling neural operators to learn solution operators from mathematically controlled spaces without inheriting discretization errors or numerical biases.

\section{Methods}
\label{sec:methods}

\subsection{Problem formulation}
\label{subsec:problem}

\begin{figure}[h!]
\centering
\includegraphics[width=1.0\textwidth]{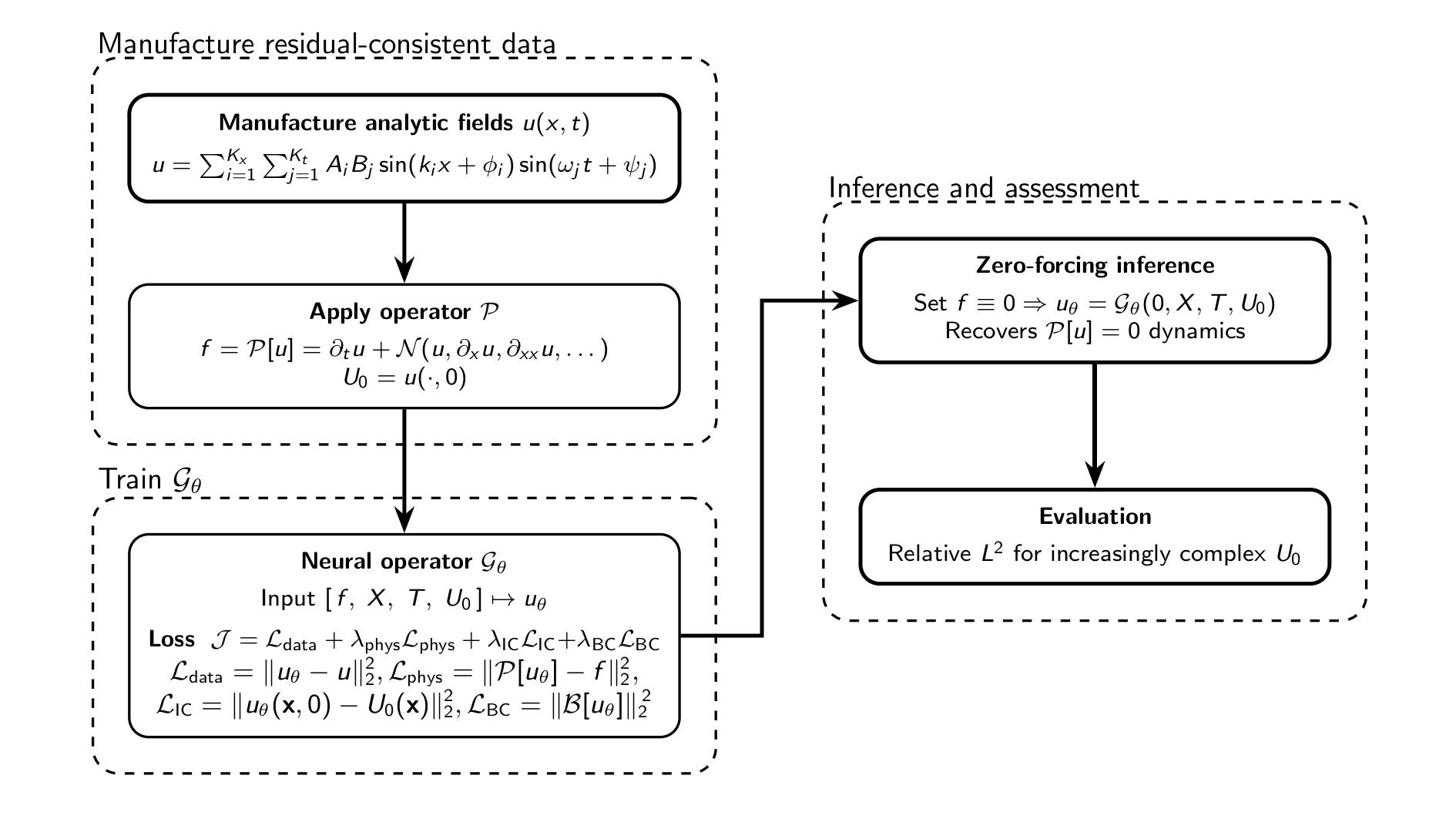}
\caption{Method of Manufactured Learning (MML) workflow.
MML constructs a self-supervised training framework by synthesizing solutions and residuals directly from the governing equations. A set of candidate functions is first introduced to automatically satisfy the prescribed initial and boundary conditions. These functions are then substituted into the governing equations, producing an analytically generated manufactured forcing term that defines the residual. This yields paired data without requiring simulations or experiments. A neural network is subsequently trained to learn the operator that maps residuals to solutions. Once trained, the manufactured forcing is set to zero, allowing the network to recover the desired physical solution of the original problem. In this way, MML removes the need for expensive pre-generated simulation or experimental data and provides a general pathway for training neural operators directly from physics.}
\label{fig:MML}
\end{figure}

We consider a general time-dependent partial differential equation (PDE)
\begin{equation}
\mathcal{P}[u] = \partial_t u + \mathcal{N}\!\big(u, \nabla u, \nabla^2 u, \dots\big) = 0,
\label{eq:pde}
\end{equation}
where \(u(\mathbf{x}, t)\) is defined on the spatio-temporal domain
\[
\Omega = \{(\mathbf{x}, t) : \mathbf{x} \in \mathcal{D} \subset \mathbb{R}^d,\; t \in [0,T]\},
\]
\(\mathcal{N}\) is a spatial differential operator, and \(\mathcal{P}\) is the full evolution operator. 
Given an initial condition \(u_0 = u(\mathbf{x}, 0)\) and appropriate boundary conditions, Eq.~\eqref{eq:pde} defines a solution operator
\begin{equation}
\mathcal{G}: u_0 \longmapsto u(\mathbf{x}, t).
\label{eq:operator}
\end{equation}
The objective of operator learning is to construct a parametric surrogate \(\mathcal{G}_\theta\) such that
\[
\mathcal{G}_\theta(u_0) \approx \mathcal{G}(u_0)
\quad \text{for all } u_0 \in \mathcal{U},
\]
where accuracy is typically measured in the \(L^2(\Omega)\) norm. 
Conventional datasets for training \(\mathcal{G}_\theta\) are obtained from repeated numerical solutions of Eq.~\eqref{eq:pde} or from experimental measurements, which are costly and often restricted to limited parameter regimes. 
To overcome these limitations, as illustrated in Fig.~\ref{fig:MML}, the MML provides a solver-free approach for constructing physics-consistent training data.

\subsection{Manufacturing residual-consistent training data}
\label{subsec:mml_data}
 
We begin by sampling smooth functions
\[
u^{(i)}(\mathbf{x}, t) \in \mathcal{U}_{\text{trial}},
\]
drawn from controlled functional spaces to generate diverse yet differentiable spatio-temporal fields. For each manufactured field \(u^{(i)}\), we compute the exact PDE residual
\begin{equation}
f^{(i)}(\mathbf{x}, t) = \mathcal{P}[u^{(i)}]
= \partial_t u^{(i)} + \mathcal{N}\!\big(u^{(i)}, \nabla u^{(i)}, \nabla^2 u^{(i)}, \dots\big),
\label{eq:residual}
\end{equation}
yielding self-consistent triplets
\[
\big(f^{(i)}, u_0^{(i)}, u^{(i)}\big),
\]
where \(u_0^{(i)} = u^{(i)}(\mathbf{x},0)\). 
Both temporal and spatial derivatives can be obtained analytically. Depending on the problem, temporal derivatives in a semi-discrete form may also be evaluated using finite differences, while spatial derivatives are computed in Fourier space under periodic boundary conditions.
In the one-dimensional setting used for our experiments, we specify manufactured fields using a separable Fourier expansion:
\begin{equation}
u(x,t) = 
\sum_{i=1}^{K_x}\sum_{j=1}^{K_t}
A_i B_j\,
\sin(k_i x + \phi_i)\,
\sin(\omega_j t + \psi_j),
\label{eq:manufactured_field}
\end{equation}
where spatial wavenumbers \(k_i \in [1, k_{\max}]\), temporal frequencies \(\omega_j \in [1, \omega_{\max}]\), amplitudes \(A_i\sim\mathcal{U}(a_{\min},a_{\max})\), \(B_j\sim\mathcal{U}(b_{\min},b_{\max})\), and phases \(\phi_i,\psi_j\sim\mathcal{U}(0,2\pi)\) ensure sample diversity. 
This representation guarantees differentiability and yields compact spectral support for higher-order derivatives. For each field \(u(x,t)\), the manufactured forcing is defined as
\begin{equation}
f(x,t) = \mathcal{P}[u(x,t)],
\label{eq:manufactured_forcing}
\end{equation}
which quantifies how strongly the trial function deviates from being an exact solution of the PDE. 
When \(f \equiv 0\), the manufactured field satisfies Eq.~\eqref{eq:pde} exactly. Each dataset instance is encoded as
\begin{equation}
\big(f, X, T, U_0\big) \longmapsto u,
\label{eq:training_tuple}
\end{equation}
where \(X(x,t)\) and \(T(x,t)\) are spatial and temporal coordinate grids, and \(U_0(x) = u(x,0)\) is broadcast in time for dimensional consistency. 
The resulting input-output pairs are stored as tensors of shape \((4,N_t,N_x)\) for the input channels \((f,X,T,U_0)\) and \((1,N_t,N_x)\) for the target field \(u\). 
These pairs form a fully analytical, solver-free training dataset.

\subsection{Neural operator training}
\label{subsec:training}

The MML framework is architecture-agnostic and can be paired with any operator-learning model.  In this work, we adopt the Fourier Neural Operator (FNO) approach \cite{li2020fourier1} as a representative example. FNO performs convolutions directly in the Fourier domain, enabling highly efficient capture of global interactions and long-range dependencies that naturally arise in the solution manifolds of PDEs. The architecture has also accumulated substantial empirical evidence of strong performance across diverse PDE families, making it a robust benchmark for evaluating new operator-learning formulations. Consequently, the insights drawn from our experiments extend beyond FNO and apply broadly to any operator-learning architecture that benefits from physically consistent, analytically constructed training data. Let
\[
u_\theta = \mathcal{G}_\theta(f, X, T, U_0)
\]
denotes the neural operator prediction parameterized by weights \(\theta\).  
To ensure that the learned operator respects both the manufactured data and the structure of the governing PDE, we train the model using the composite objective
\begin{equation}
\mathcal{J}(\theta) =
\underbrace{\|u_\theta - u\|_{2}^{\,2}}_{\mathcal{L}_{\text{data}}}
+
\lambda_{\text{phys}}
\underbrace{\|\mathcal{P}[u_\theta] - f\|_{2}^{\,2}}_{\mathcal{L}_{\text{phys}}}
+
\lambda_{\text{IC}}
\underbrace{\|u_\theta(\mathbf{x},0) - U_0(\mathbf{x})\|_{2}^{\,2}}_{\mathcal{L}_{\text{IC}}}
+
\lambda_{\text{BC}}
\underbrace{\|\mathcal{B}[u_\theta]\|_{2}^{\,2}}_{\mathcal{L}_{\text{BC}}},
\label{eq:loss_full}
\end{equation}
where \(\mathcal{L}_{\text{data}}\) enforces agreement with the manufactured solutions,  
\(\mathcal{L}_{\text{phys}}\) penalizes mismatch between the predicted residual and the manufactured forcing,  
and \(\mathcal{L}_{\text{IC}}\) and \(\mathcal{L}_{\text{BC}}\) ensure consistency with the prescribed initial and boundary conditions, respectively.  

Because all PDEs studied here are posed on periodic spatial domains, the boundary-consistency term \(\mathcal{L}_{\text{BC}}\) reduces to enforcing periodicity via the operator
\[
\mathcal{B}[u_\theta] = u_\theta(\mathbf{x}_{\min}, t) - u_\theta(\mathbf{x}_{\max}, t),
\]
ensuring that the predicted solution matches exactly at the domain endpoints for all times.  
The weighting coefficients \(\lambda_{\text{phys}}, \lambda_{\text{IC}}, \lambda_{\text{BC}}\) regulate the balance between data fidelity, PDE consistency, and periodic boundary enforcement.

Through this residual and constraint aware training procedure, the neural operator learns mappings that remain faithful to the underlying differential operator \(\mathcal{P}\), even when trained solely on analytically manufactured forcing fields.  
After training, we perform zero-forcing inference by setting \(f \equiv 0\), yielding
\begin{equation}
\mathcal{P}[u_\theta] = 0
\quad \Longrightarrow \quad
u_\theta = \mathcal{G}_\theta(f=0, X, T, U_0),
\label{eq:zero_forcing_inference}
\end{equation}
which allows the operator to recover the intrinsic unforced dynamics of the original PDE.

\subsection{Model evaluation and generalization}
\label{subsec:evaluation}

The performance of neural operators trained under the MML framework is evaluated using quantitative error metrics applied consistently across all PDE classes. Model accuracy is assessed using the relative \(L^2\) error,
\begin{equation}
\epsilon_{L^2}
= \frac{\|u_\theta - u\|_{L^2(\Omega)}}{\|u\|_{L^2(\Omega)}},
\label{eq:rel_l2_metric}
\end{equation}
which provides a global measure of agreement between the predicted and reference fields. To examine the model’s ability to generalize beyond the manufactured training manifold, we employ a unified zero-forcing protocol in which the forcing is set to \(f \equiv 0\) during inference. The trained operator is then evaluated on a hierarchy of initial conditions with progressively richer spectral content and nonlinear interactions, none of which were used for training. These include a single sinusoidal mode, a two-mode superposition, and a three-mode configuration with phase shifts and varying wavenumbers. This progression enables a systematic assessment of the operator’s capacity to reproduce more complex dynamical structures, capture higher-order interactions, and maintain stability across broadband initial spectra.

For each initial condition, reference solutions are generated either analytically or via high-resolution numerical solvers. Comparisons between the learned and reference trajectories provide a stringent test of the operator’s ability to recover physically correct zero-forcing dynamics. This evaluation strategy ensures that model performance is judged not only by its agreement with manufactured training data but also by its ability to generalize to unseen, physically meaningful scenarios, thereby validating the core premise of the MML framework.

\section{Results}

We evaluate the Method of Manufactured Learning (MML) across four canonical time-dependent PDEs: the heat equation, the linear advection equation, the viscous Burgers equation, and a nonlinear diffusion–reaction equation.
All experiments employ the same FNO-based neural operator architecture and training protocol, differing only in the choice of the governing operator \(\mathcal{P}[u]\). This uniformity isolates the contribution of MML from architectural factors and enables a direct comparison of performance across PDE families.

For each PDE, we discretize the spatio-temporal domain using \(N_x = 128\) spatial points and \(N_t = 128\) time levels, and construct a manufactured dataset comprising \(1024\) training samples and \(32\) validation samples. An FNO-based neural operator \(\mathcal{G}_\theta\) is trained on these datasets using the composite objective in Eq.~\eqref{eq:loss_full}. All models share the same architecture, consisting of four Fourier layers with \(64\) channels and \(40\) retained modes in both time and space, so that observed differences in accuracy and generalization can be attributed to the PDE-specific operator structure and the MML-generated data rather than to changes in model capacity.

\subsection{Heat equation}
\label{subsec:heat_results}

We begin by evaluating the MML on the canonical one-dimensional heat equation
\begin{equation}
u_t - \nu u_{xx} = 0,
\quad x \in [0,2\pi], \quad t \in [0,1],
\label{eq:heat_pde}
\end{equation}
with periodic boundary conditions and viscosity $\nu = 2\times 10^{-1}$.  
To assess whether an operator trained exclusively on manufactured data can recover the true homogeneous dynamics, we test three increasingly complex initial conditions:
\begin{align}
u_0^{(1)}(x) &= 0.8\,\sin(x), \label{eq:heat_ic1}\\
u_0^{(2)}(x) &= 0.5\,\sin(x) - 0.8\,\sin(3x + 0.7), \label{eq:heat_ic2}\\
u_0^{(3)}(x) &= 0.9\,\sin(x) - 0.3\,\sin(3x + 0.7) + 0.7\,\sin(5x - 1.2), \label{eq:heat_ic3}
\end{align}
corresponding to single-mode, two-mode, and three-mode Fourier superpositions.  
Reference solutions are obtained analytically using the exact spectral decay
\begin{equation}
\widehat{u}(k,t) = \widehat{u}(k,0)\,\exp(-\nu k^2 t),
\label{eq:heat_exact_spectral}
\end{equation}
and compared against the zero-forcing predictions 
\(
u_\theta = \mathcal{G}_\theta(f=0,X,T,U_0).
\)
\begin{figure}[h!]
\centering
\includegraphics[width=\textwidth]{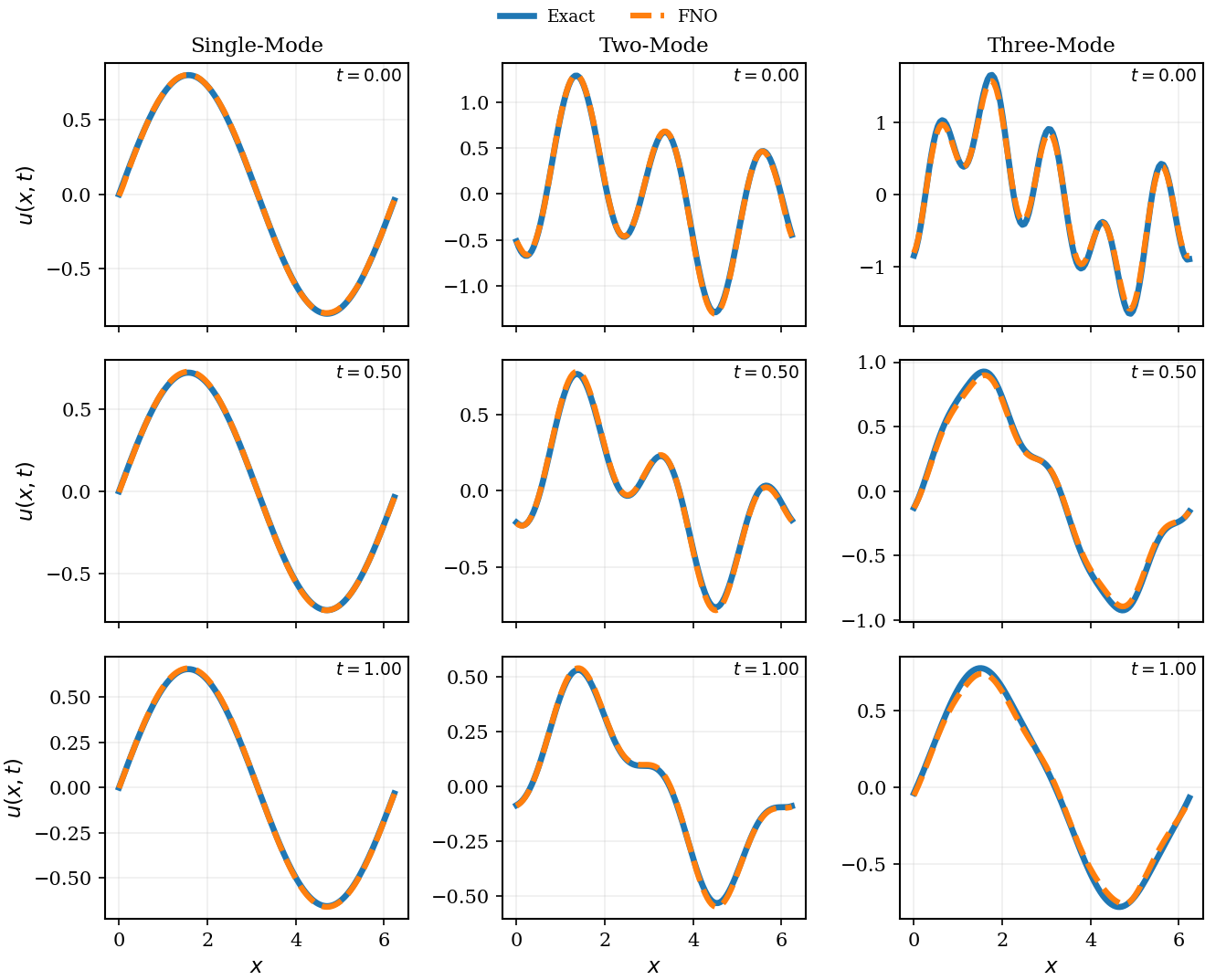}
\caption{Time-slice comparisons for the one-dimensional heat equation under three unseen initial conditions.
Columns correspond to the single-mode, two-mode, and three-mode initial configurations, while rows show the solution profiles at 
$t = 0$, $t = 0.5$, and $t = 1.0$. Higher-frequency modes are smoothed at the correct rate, and the predicted trajectories maintain spatial periodicity and temporal coherence throughout the full evolution.}
\label{fig:heat_timeslices}
\end{figure}
Figure~\ref{fig:heat_timeslices} presents time-slice comparisons at $t=0$, $t=0.5$, and $t=1.0$.  
Across all initial conditions, the MML-trained operator reproduces the correct diffusive decay with near-perfect overlap between the predicted and exact solutions.  
Even for the broadband three-mode configuration, the model accurately captures both the rapid damping of high-frequency components and the progressive smoothing of the solution over time.
\begin{figure}[h!]
\centering
\includegraphics[width=\textwidth]{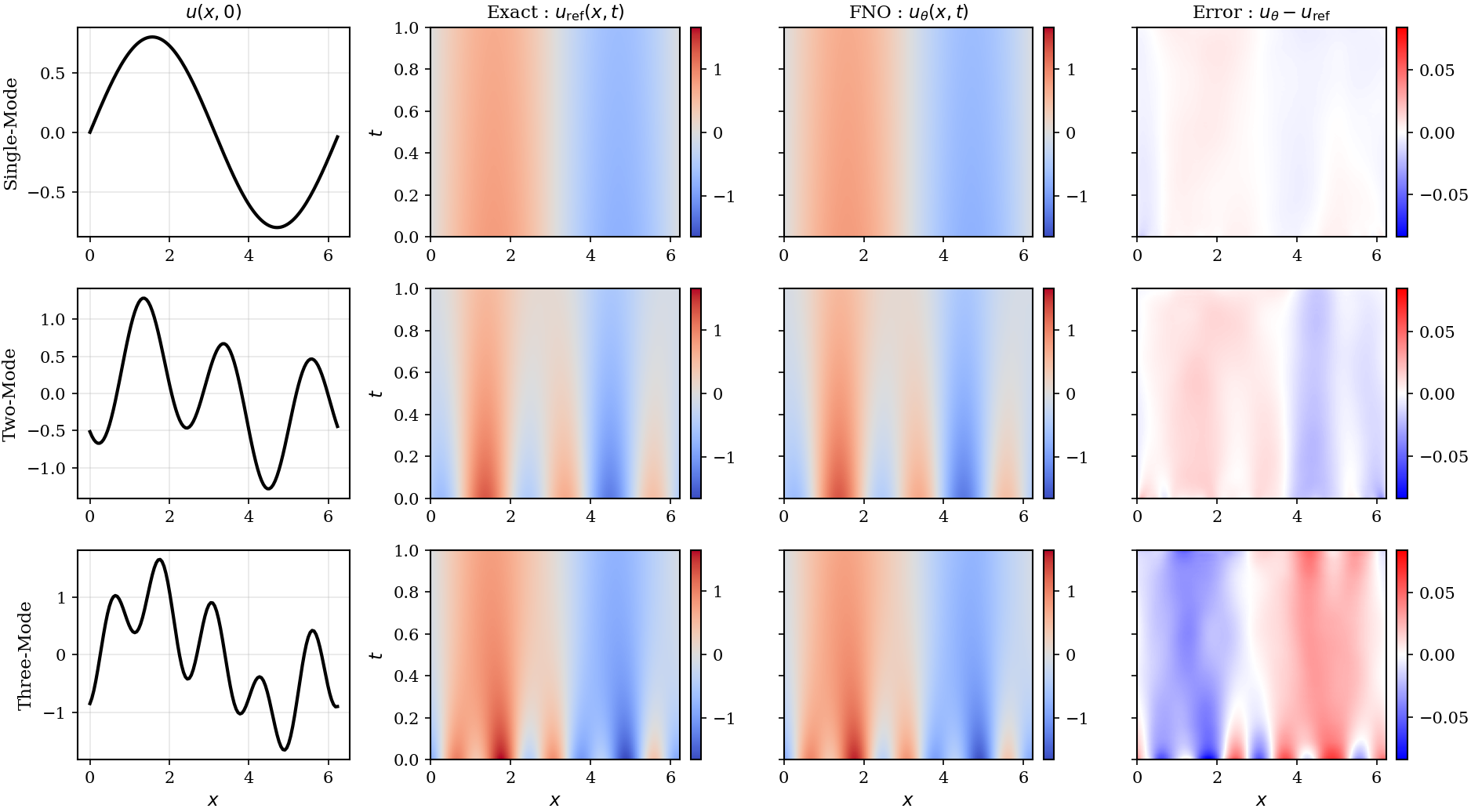}
\caption{
Spatio–temporal evolution of the one-dimensional heat equation under three unseen initial conditions. The corresponding relative $L^2$ errors for the predicted solution fields are  
$0.7158\%$ (single-mode),  
$2.361\%$ (two-mode),  
and $3.886\%$ (three-mode),  
demonstrating high quantitative accuracy and robust generalization to broadband initial conditions.
}
\label{fig:heat_fields}
\end{figure}
The full spatio-temporal fields shown in Fig.~\ref{fig:heat_fields} further illustrate the fidelity of the learned operator.  
The predicted heatmaps closely match the exact diffusive evolution, preserving spatial periodicity and exhibiting accurate attenuation of all resolved modes. The discrepancy between the FNO predictions and the analytical solution remains small throughout the entire domain, with errors concentrated only in regions where high-frequency components decay rapidly. Even in the most broadband three-mode case, the deviation between predicted and reference fields stays smooth, low in magnitude, and free of spurious oscillations, indicating that the operator captures both the correct dissipation rate and the global energy decay.  

\subsection{Advection equation}
\label{subsec:advection_results}

We next evaluate the MML on the one-dimensional linear advection equation
\begin{equation}
u_t + c\,u_x = 0, \qquad c = 0.5,
\quad x \in [0,2\pi], \quad t \in [0,1],
\label{eq:advection_pde}
\end{equation}
posed with periodic boundary conditions.  
Unlike the diffusive dynamics of the heat equation, the advection equation transports the initial condition without deformation, making it a stringent test of phase accuracy and long-time stability. To probe generalization, we consider the same hierarchy of initial conditions as in the heat equation~\ref{subsec:heat_results}. Reference solutions are computed exactly using the spectral phase shift
\begin{equation}
\widehat{u}(k,t) = \widehat{u}(k,0)\,e^{-\,i c k t},
\label{eq:adv_exact_spectral}
\end{equation}
and compared against zero-forcing predictions 
\(
u_\theta = \mathcal{G}_\theta(f=0,X,T,U_0).
\)

\begin{figure}[h!]
\centering
\includegraphics[width=\textwidth]{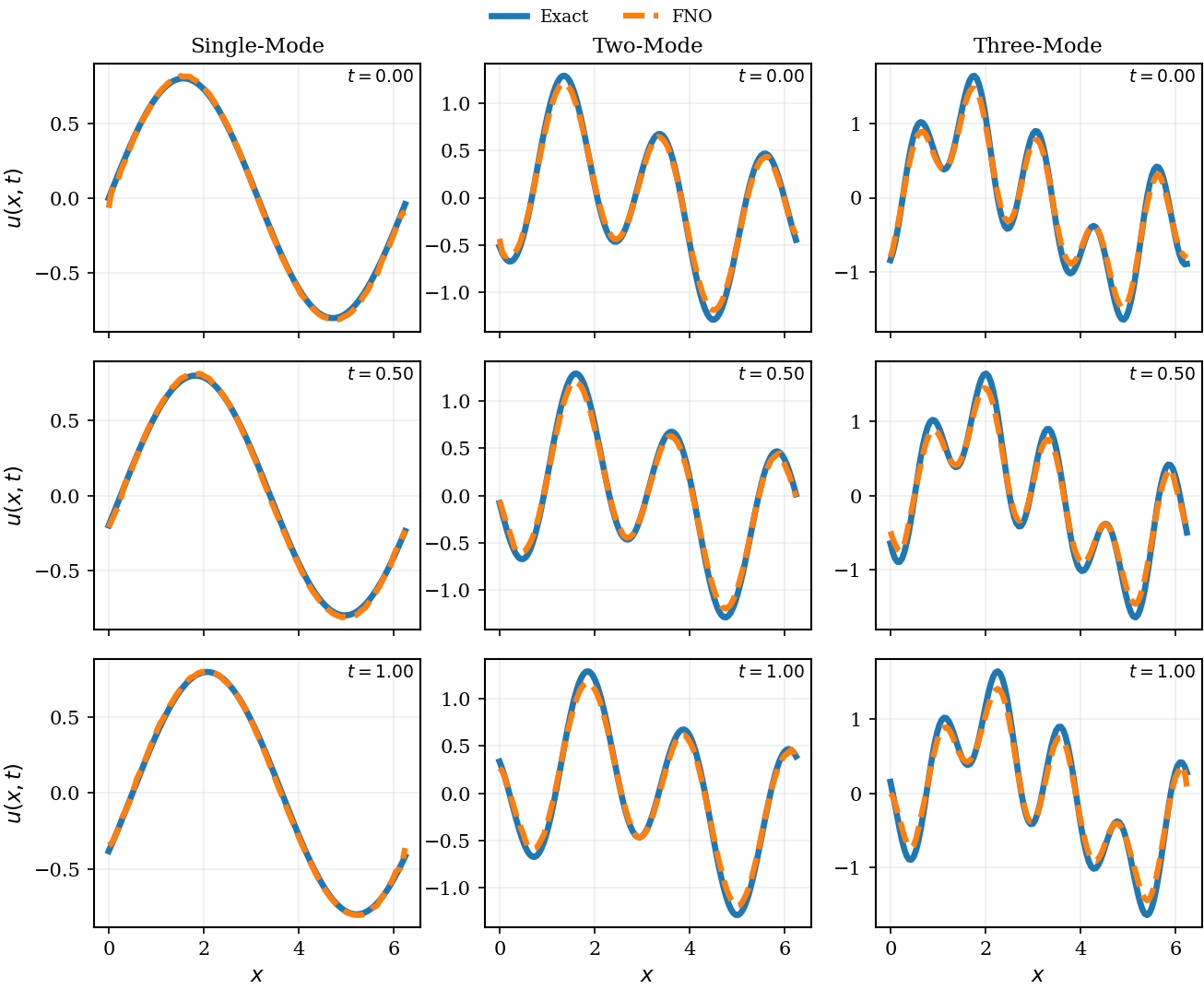}
\caption{
Time-slice comparisons for the linear advection equation under three unseen initial conditions.
Each column corresponds to a different initial condition, and rows show the solution at 
$t = 0$, $t = 0.5$, and $t = 1$.
The MML-trained operator accurately captures the advective transport, preserving wave shapes and phase information across the entire time horizon.}
\label{fig:adv_timeslices}
\end{figure}

Figure~\ref{fig:adv_timeslices} shows the resulting time-slice comparisons.  
Across all initial conditions, the MML-trained operator correctly reproduces the characteristic translation of the waveform at constant speed.  
The predicted trajectories exhibit minimal phase drift, even for the three-mode broadband input where multiple wavenumbers must propagate coherently.  
Unlike diffusive systems, advection requires the operator to preserve fine-scale structure without attenuation, a property recovered by the MML-trained model.

\begin{figure}[h!]
\centering
\includegraphics[width=\textwidth]{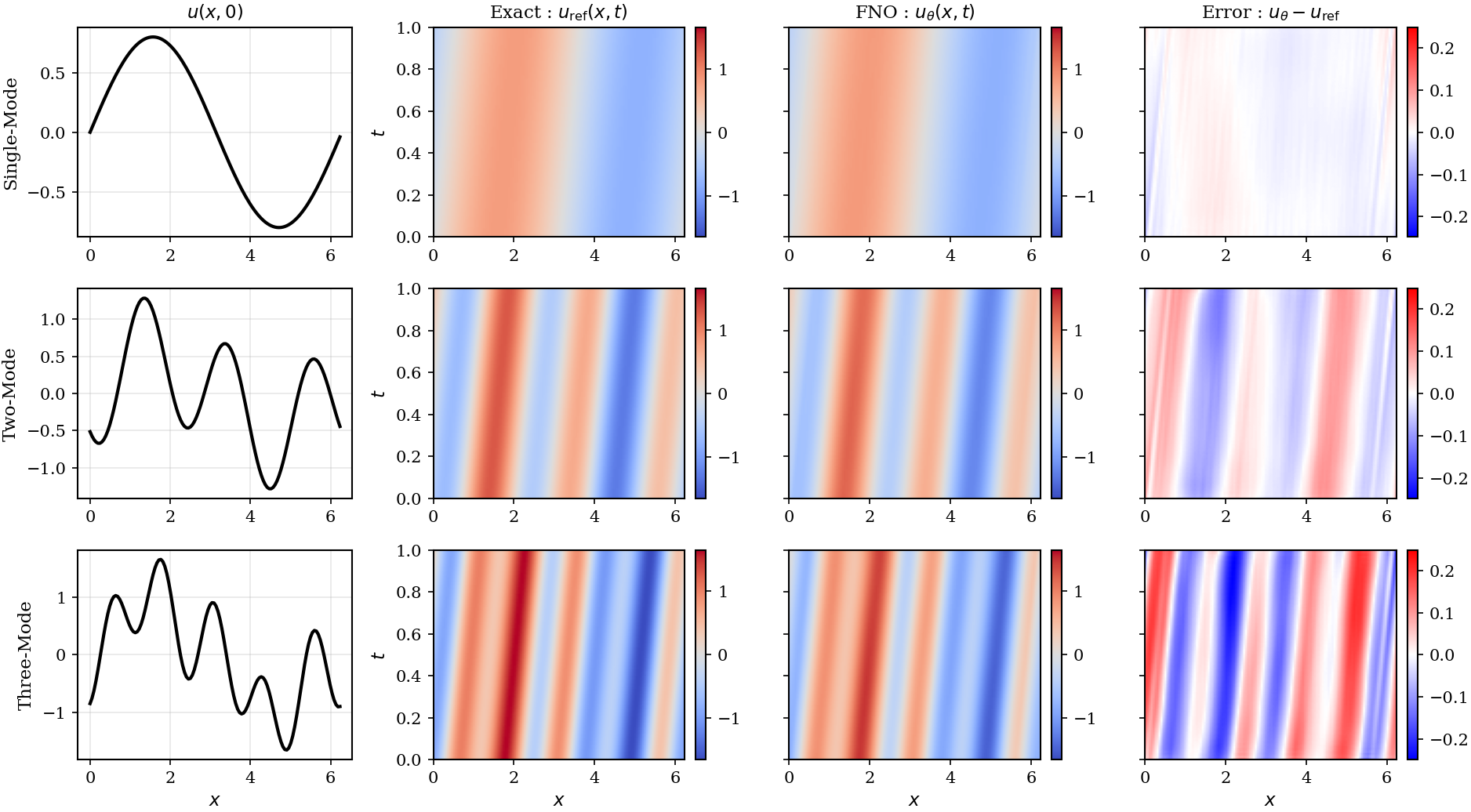}
\caption{
Spatio-temporal evolution of the one-dimensional advection equation predicted by the MML-trained operator for three unseen initial conditions.
The operator faithfully reconstructs the downstream transport without artificial diffusion or distortion.  
The corresponding relative $L^2$ errors are
$2.14\%$ (single-mode),
$7.994\%$ (two-mode),
and $12.37\%$ (three-mode),
demonstrating accurate yet increasingly challenging reconstruction as spectral complexity increases.}
\label{fig:adv_fields}
\end{figure}

\begin{figure}[h!]
\centering
\includegraphics[width=0.6\textwidth]{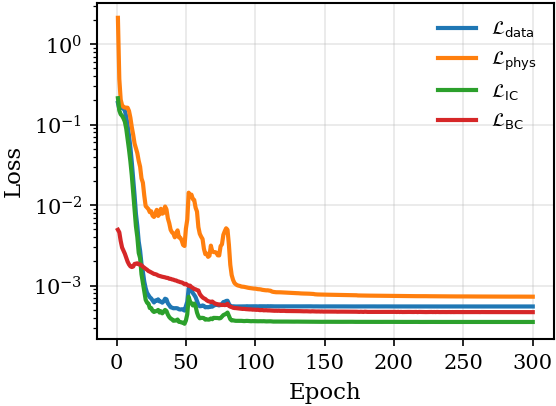}
\caption{
Training history for the one-dimensional advection equation under the Method of Manufactured Learning.  
The curves show the evolution of the data loss $\mathcal{L}_{\text{data}}$, physics residual loss $\mathcal{L}_{\text{phys}}$, initial-condition loss $\mathcal{L}_{\text{IC}}$, and periodic boundary-consistency loss $\mathcal{L}_{\text{BC}}$ over 300 epochs.  
All loss components decrease monotonically and settle into a stable plateau, indicating that the FNO successfully balances data fidelity, PDE consistency, and constraint enforcement throughout training.
}
\label{fig:adv_training_history}
\end{figure}

Figure~\ref{fig:adv_training_history} summarizes the evolution of all loss components during MML training for the advection equation. The full spatio-temporal fields in Fig.~\ref{fig:adv_fields} confirm that the learned operator reproduces the correct advective transport across the domain.  
The waveforms remain sharp, the periodic boundary is respected, and the model avoids artificial numerical dissipation despite never observing true advective rollouts during training.  The accompanying error fields reveal that the discrepancies between the predicted and exact solutions remain structured and low in magnitude, with errors primarily concentrated along regions of steep phase propagation.  
For the single-mode configuration, the residuals stay uniformly small, indicating near perfect phase alignment over the entire time horizon.  
As the spectral content of the initial condition increases, the error field exhibits coherent oscillatory bands reflecting slight phase shifts rather than amplitude distortion or spurious errors.  
Importantly, no unphysical growth or instability is observed, and the error remains bounded and smooth throughout space-time, demonstrating that the MML-trained operator successfully captures the pure transport dynamics and preserves waveform integrity even in the broadband three-mode regime.

\subsection{Burgers equation}
\label{subsec:Burgers_results}

We next examine the one-dimensional viscous Burgers equation
\begin{equation}
u_t + u\,u_x - \nu u_{xx} = 0,
\quad x \in [0,2\pi], \quad t \in [0,1],
\label{eq:Burgers_pde}
\end{equation}
with periodic boundary conditions and viscosity $\nu = 5\times 10^{-2}$, chosen to keep the solution smooth over the time interval of interest. We assess zero-forcing generalization using 
\begin{align}
u_0^{(1)}(x) &= 0.8\,\sin(x), \label{eq:Burgers_ic1}\\
u_0^{(2)}(x) &= 0.8\,\sin(x) - 0.3\,\sin(3x + 0.7), \label{eq:Burgers_ic2}\\
u_0^{(3)}(x) &= -0.8\,\sin(x) + 0.3\,\sin(3x + 0.7) - 0.2\,\sin(5x - 1.1), \label{eq:Burgers_ic3}
\end{align}
Reference trajectories are generated using a Fourier pseudo-spectral solver with sufficiently fine temporal and spatial resolution to resolve shock like steepening and viscous smoothing.

\begin{figure}[h!]
\centering
\includegraphics[width=0.9\textwidth]{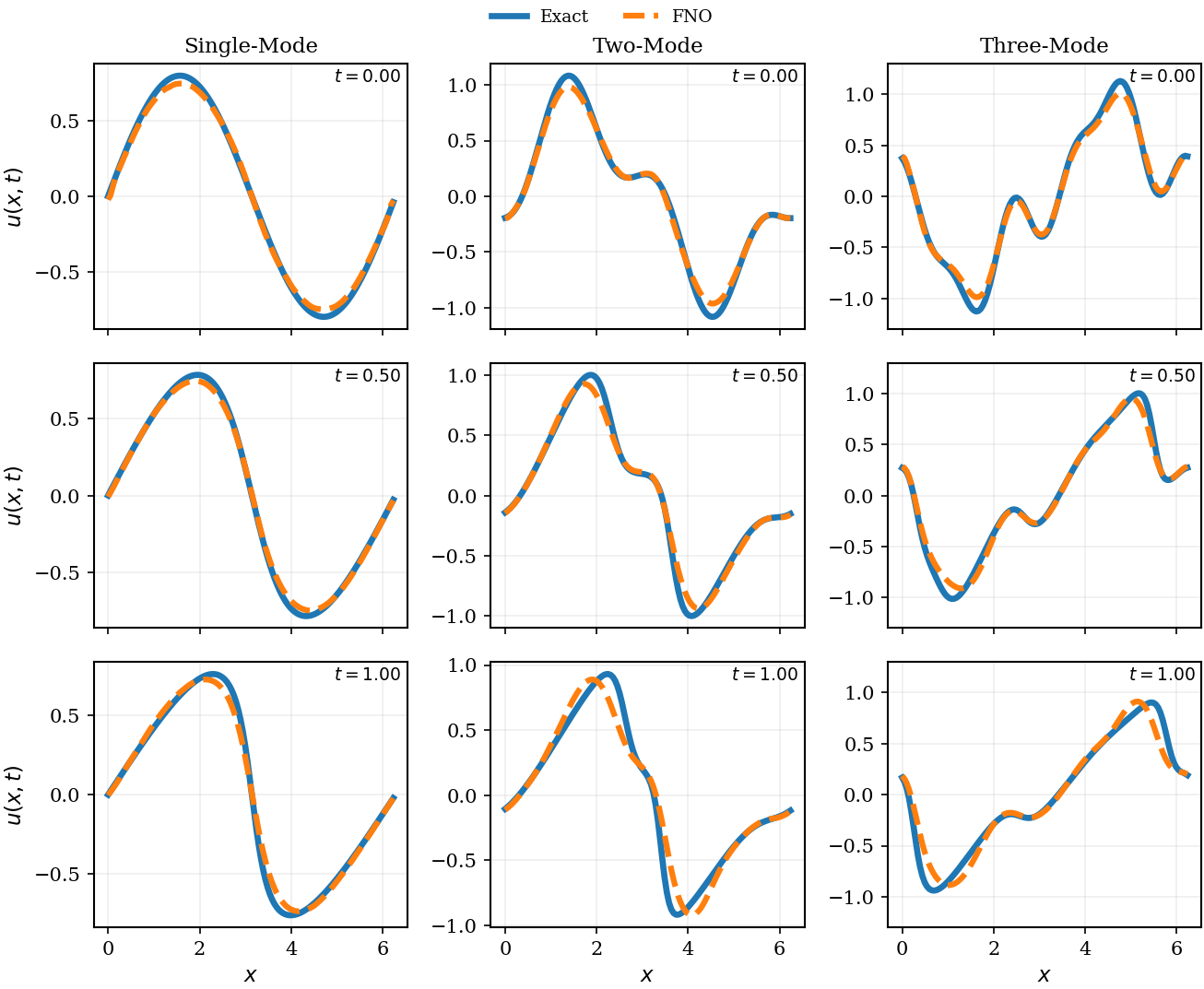}
\caption{
Time-slice comparisons for the one-dimensional viscous Burgers equation under single, two, and three-mode initial conditions.  The trained FNO (dashed curves) closely tracks the reference solution (solid curves) as wavefronts advect, steepen, and subsequently smooth under viscous dissipation, demonstrating that the learned operator faithfully reproduces the nonlinear transport-diffusion balance in Eq.~\eqref{eq:Burgers_pde} for all tested initial data.
}
\label{fig:Burgers_timeslices}
\end{figure}
Figure~\ref{fig:Burgers_timeslices} shows time-slice comparisons for all three initial conditions.  
For the single-mode case, the FNO prediction is visually indistinguishable from the reference solution at all times, capturing both the mild steepening and gradual decay in amplitude.  In the two-mode and three-mode cases, the model continues to track the formation and advection of sharper gradients as nonlinear mode coupling redistributes energy across scales.  
The predicted profiles exhibit no spurious oscillations near steep fronts.
\begin{figure}[h!]
\centering
\includegraphics[width=0.9\textwidth]{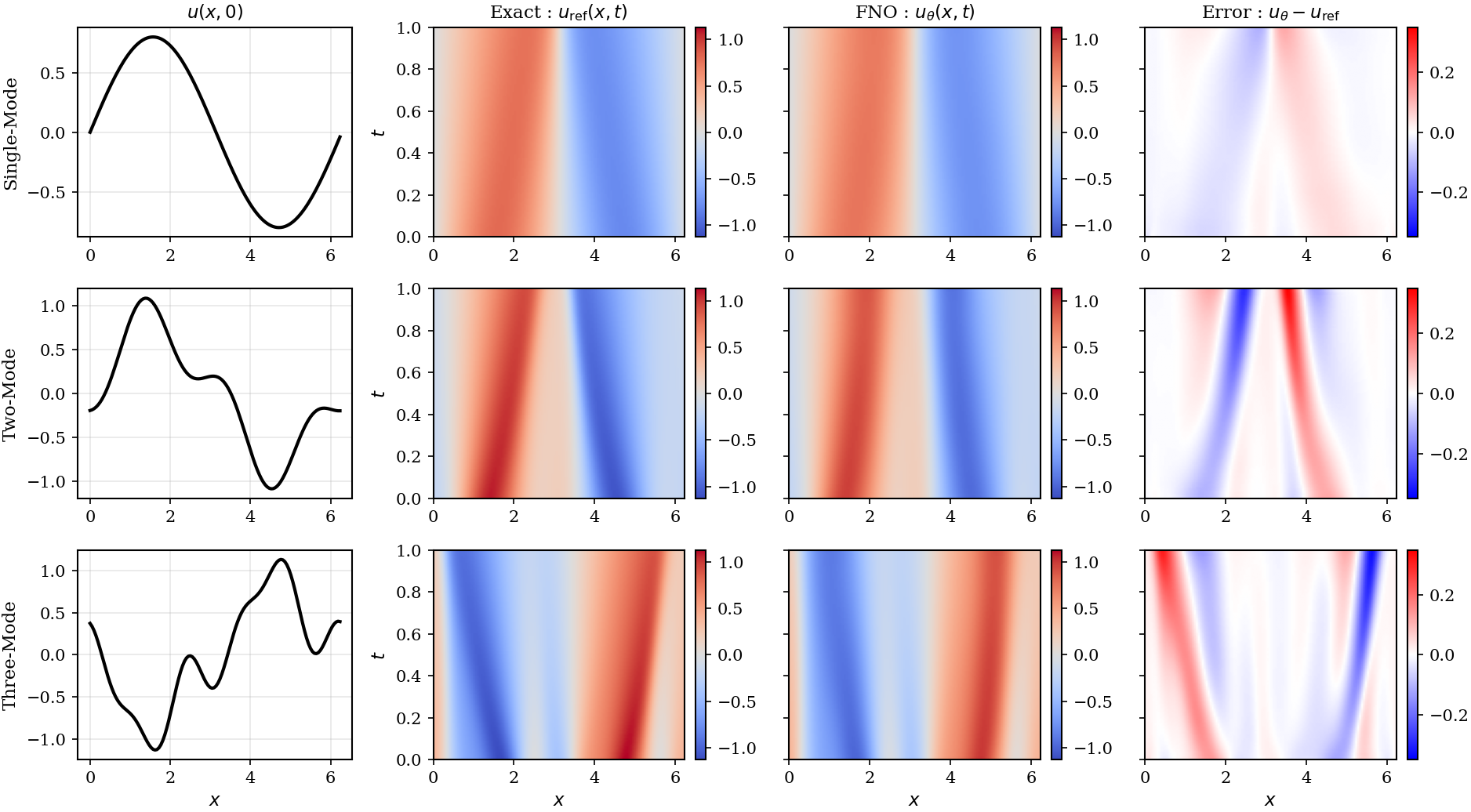}
\caption{
Spatio-temporal evolution of the viscous Burgers equation under three unseen initial conditions.  
The corresponding relative $L^2$ errors are  
$5.447\times10^{-2}$ for the single-mode initial condition,  
$1.189\times10^{-1}$ for the two-mode case, and  
$1.259\times10^{-1}$ for the three-mode case.
}
\label{fig:Burgers_fields}
\end{figure}
The full spatio-temporal fields in Fig.~\ref{fig:Burgers_fields} further highlight the ability of MML to train a neural operator for nonlinear dynamics.  
The FNO reconstructions preserve the phase and amplitude of the dominant structures over the entire time horizon, while the error fields remain small and structured.

\subsection{Diffusion-reaction equation}
\label{subsec:dr_results}

We conclude our evaluation of the MML with the one-dimensional diffusion-reaction equation
\begin{equation}
u_t = \nu\,u_{xx} - u + u^3,
\qquad x \in [0,2\pi], \quad t \in [0,1],
\label{eq:dr_pde}
\end{equation}
posed with periodic boundary conditions and viscosity \(\nu = 5\times10^{-2}\). 
This nonlinear equation exhibits the competing effects of diffusion, linear damping, and cubic self-activation.  
As in the previous test cases, the neural operator is trained solely on analytically manufactured fields and evaluated under zero forcing (\(f \equiv 0\)) to assess whether the learned operator can recover the intrinsic dynamics of Eq.~\eqref{eq:dr_pde}.

To probe generalization from manufactured data to true physical evolution, we again consider three increasingly complex Fourier-mode initial conditions:
\begin{align}
u_0^{(1)}(x) &= 0.8\,\sin(x), \label{eq:dr_ic1}\\[3pt]
u_0^{(2)}(x) &= -0.5\,\sin(x) + 0.8\,\sin(3x + 0.7), \label{eq:dr_ic2}\\[3pt]
u_0^{(3)}(x) &= 0.9\,\sin(x) - 0.3\,\sin(3x + 0.7) + 0.7\,\sin(5x - 1.2), \label{eq:dr_ic3}
\end{align}
representing single-mode, two-mode, and broadband three-mode configurations, respectively. Reference solutions are generated numerically using a high-resolution spectral integrator .
\begin{figure}[h!]
\centering
\includegraphics[width=0.92\textwidth]{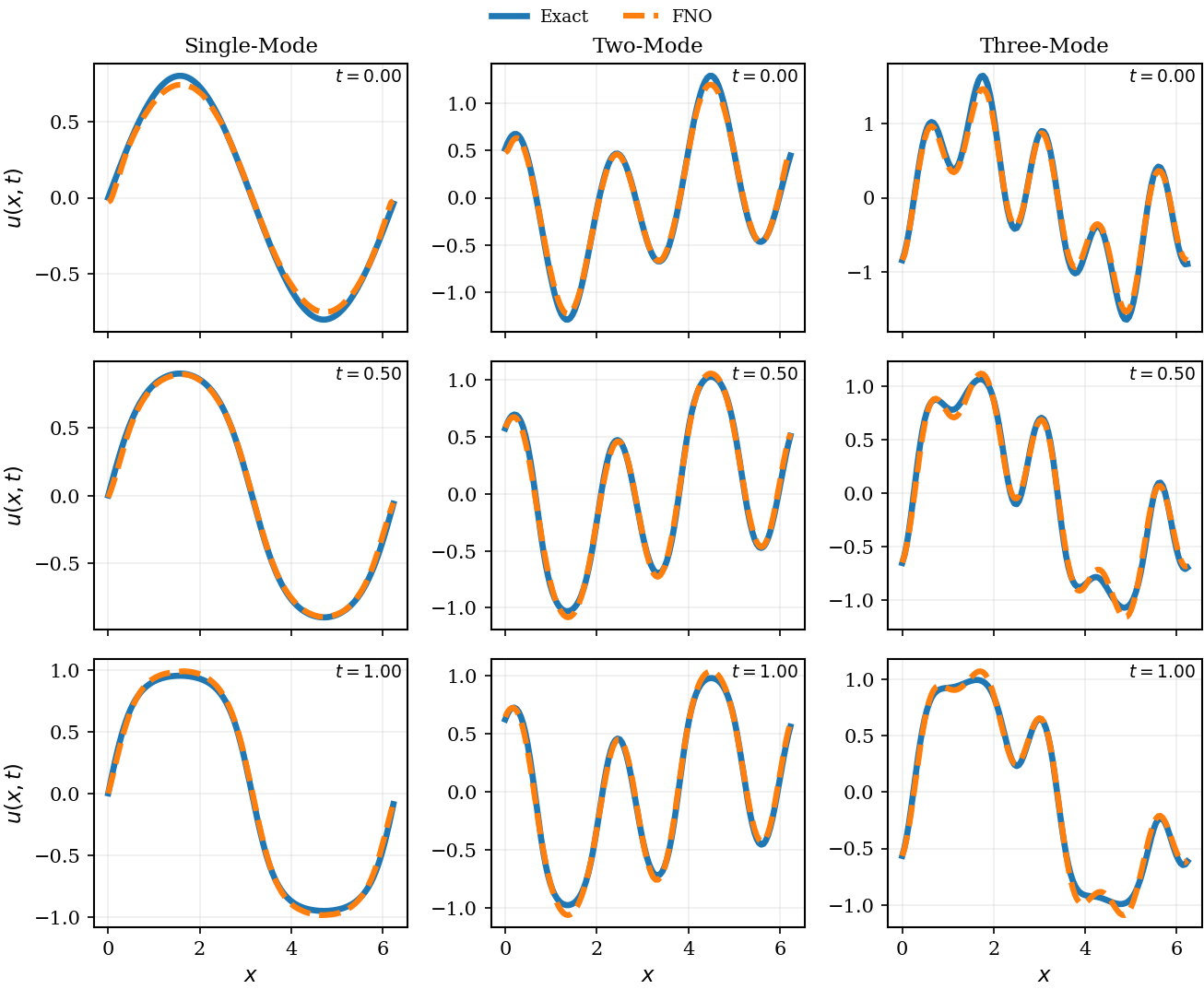}
\caption{
Time-slice comparisons for the one-dimensional diffusion-reaction equation under three unseen initial conditions.  
Across all cases, the MML-trained operator accurately tracks the nonlinear interplay between diffusion, linear decay, and cubic growth, with predicted profiles exhibiting strong agreement with the reference solution throughout the evolution window.
}
\label{fig:dr_timeslices}
\end{figure}

Figure~\ref{fig:dr_timeslices} shows the evolution of solution profiles at representative time slices.  
The MML-trained neural operator reproduces the smoothing and amplitude-modulating behaviour characteristic of diffusion-reaction systems.  
Even for the broadband three-mode initial condition, the predicted rollouts closely follow the steepness changes and redistribution of modal energy induced by the nonlinear reaction term, indicating that the operator generalizes beyond the manufactured training manifold.

\begin{figure}[h!]
\centering
\includegraphics[width=0.92\textwidth]{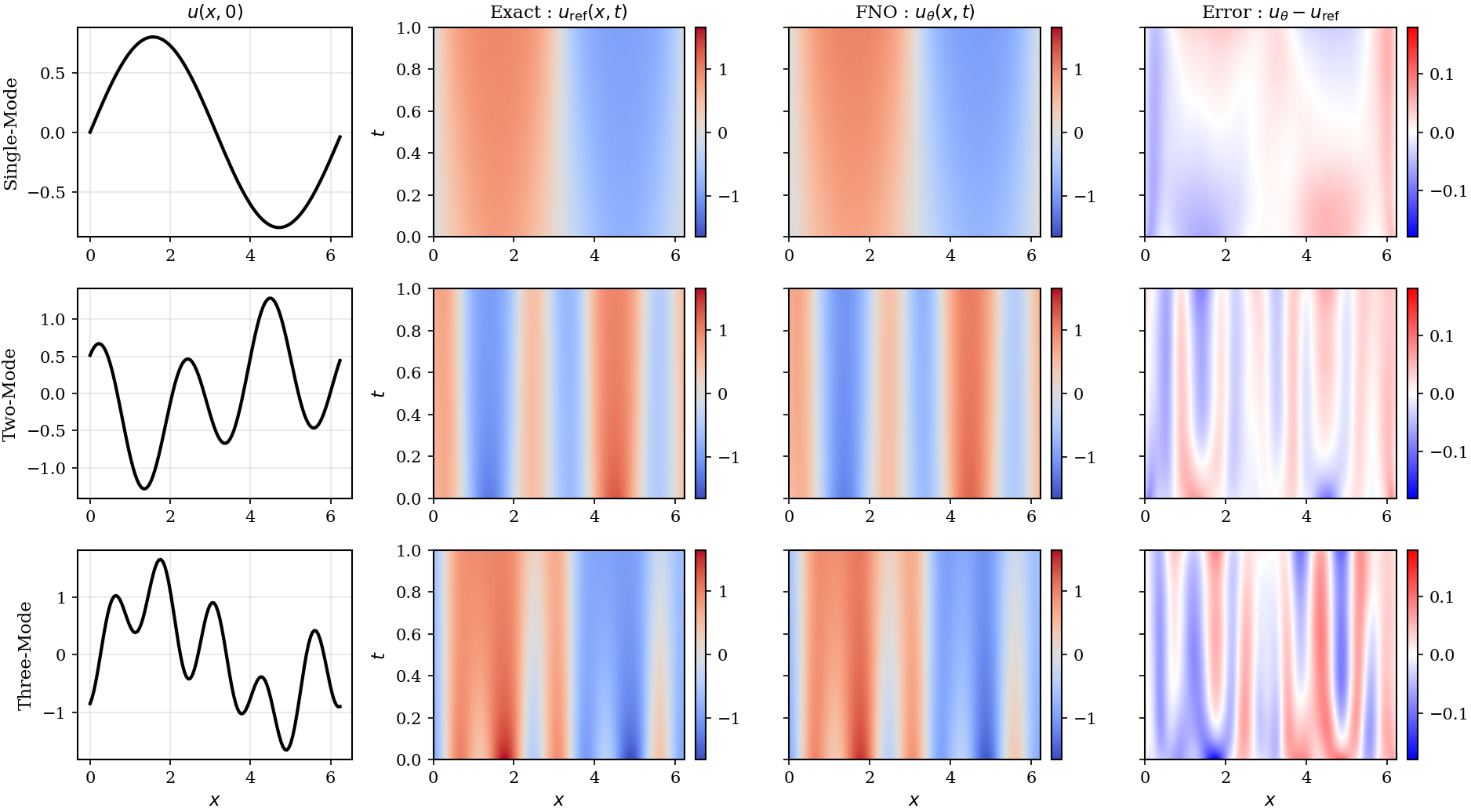}
\caption{
Spatio-temporal evolution of the diffusion-reaction equation for three unseen initial conditions.  
The relative $L^{2}$ errors of the predicted fields are  
$3.505\%$ (single-mode),  
$4.391\%$ (two-mode), and  
$5.944\%$ (three-mode),  
indicating strong quantitative agreement despite the nonlinear reaction dynamics and increasing spectral richness of the initial states.
}
\label{fig:dr_fields}
\end{figure}

The full solution fields in Fig.~\ref{fig:dr_fields} highlight the fidelity of the MML-trained operator under nonlinear dynamics.  
The predicted heatmaps correctly capture the interaction between diffusive smoothing and nonlinear growth, producing coherent spatial patterns and accurate temporal decay rates.  
Quantitatively, the relative \(L^2\) errors remain low across all three tests, increasing only moderately as the initial condition becomes more broadband, which is consistent with the greater complexity of the resulting nonlinear evolution.  
The error fields remain smooth, structured, and largely confined to regions where strong cubic interactions dominate, signalling that the model captures the principal dynamics while exhibiting only mild phase and amplitude discrepancies.  
Overall, these results demonstrate that MML enables neural operators to approximate the nonlinear behavior of the diffusion-reaction mechanism with high accuracy using only analytically manufactured training data, generated directly from Eq.~\eqref{eq:dr_pde}, and without any numerical solution of the equation before or during training.

\section{Summary and Conclusions}


This work introduced the \emph{Method of Manufactured Learning} (MML), a solver-independent framework for training neural operators using analytically constructed, physics-consistent datasets. We extend the method of manufactured solutions philosophy to scientific machine learning by formulating a solver-free framework that synthesizes analytical solution–forcing pairs for training neural operators. Inspired by the classical MMS approach, MML reframes data generation as a process of \emph{functional synthesis} rather than numerical simulation.
By prescribing smooth, diverse trial functions and computing the associated forcing fields through exact application of the governing operator, the method produces residual-consistent training data that embed the structure of the PDE directly into the learning process. This eliminates reliance on computationally intensive solvers, avoids discretization errors, and enables the creation of large, noise–free datasets with precisely controlled spectral content.

To assess the effectiveness of this paradigm, we trained a Fourier Neural Operator (FNO) on manufactured datasets for four canonical time-dependent PDEs: the heat, linear advection, viscous Burgers, and nonlinear diffusion-reaction equations. In all cases, the neural operator was trained only on manufactured data, yet recovered the correct unforced dynamics when evaluated under a zero-forcing protocol. Table~\ref{tab:all_errors} summarizes the achieved relative $L^2$ errors across the three increasingly complex initial conditions used for each PDE.

\begin{table}[h!]
\centering
\caption{Summary of relative $L^2$ errors for all PDEs and all initial-condition complexities under zero-forcing inference.}
\vspace{0.4em}
\label{tab:all_errors}
\begin{tabular}{lccc}
\toprule
\textbf{PDE} & \textbf{Single–mode} & \textbf{Two–mode} & \textbf{Three–mode} \\
\midrule
Heat & $0.7158\%$ & $2.361\%$ & $3.886\%$ \\
Linear Advection & $2.14\%$ & $7.994\%$ & $12.37\%$ \\
Viscous Burgers & $5.447\%$ & $11.89\%$ & $12.59\%$ \\
Diffusion–Reaction & $3.505\%$ & $4.391\%$ & $5.944\%$ \\
\bottomrule
\end{tabular}
\end{table}

Across all four PDE families, the MML-trained operator demonstrated strong quantitative accuracy and robust qualitative behavior. The heat equation exhibited the lowest errors, consistent with its linear diffusive dynamics. Linear advection produced larger errors due to long-time transport sensitivity, which the operator nonetheless captured without artificial diffusion. Burgers equation introduced nonlinear steepening, yet the model reproduced its dissipative transport structure and remained stable. The diffusion-reaction equation, containing both nonlinear source and diffusive terms, was accurately predicted even for broadband three-mode initial conditions. These consistent results across distinct physical regimes highlight the ability of MML to endow neural operators with physically meaningful inductive bias purely through manufactured residuals.

Beyond the specific demonstrations provided here, the Method of Manufactured Learning represents a general and architecture-agnostic strategy for constructing training corpora in operator learning. Because it bypasses numerical solvers entirely, both in data generation for training and within the training process itself, MML offers a scalable avenue for pretraining large neural operators, rapidly prototyping architectural variants, and systematically characterizing operator-learning behavior under controlled functional families. As neural operators continue to expand toward high-dimensional, multi-physics, and data–scarce regimes, MML provides a mathematically principled and computationally efficient pathway for generating large, diverse, and physically grounded datasets without incurring the prohibitive cost of PDE simulations.

Future directions include extending MML to multidimensional, multi–physics, chaotic and stiff settings, incorporating irregular and nonperiodic geometries, constructing manufactured functional spaces capable of representing shocks or sharp-gradient solutions, and combining manufactured corpora with solver-generated data to further enhance generalization. In summary, the Method of Manufactured Learning offers a foundation for scalable, solver–free operator learning by embedding the governing physics directly through analytically manufactured residuals, positioning it as a promising paradigm for the next generation of scientific machine learning models.

\backmatter

\section*{Funding Declaration}
This work was supported in part by the Air Force Office of Scientific Research 
(AFOSR) under Grant No.\ FA9550\textendash 24\textendash 1\textendash 0327. 


\section*{Author Contributions}

O.\,S.\ conceived the Method of Manufactured Learning, provided theoretical guidance, and supervised the project. A.\,S.\ designed and executed the neural operator framework, performed the numerical experiments, carried out the analyses, and prepared the initial draft. Both authors discussed the results, contributed to the writing, and approved the final manuscript.


\section*{Competing Interests}

The authors declare no competing interests.

\section*{Data Availibility}

All codes, experimental findings, and trained model results associated with this 
work are publicly available in our GitHub repository: 
\url{https://github.com/as26101999/Method-of-Manufacture-Learning-for-Synthetic-Training-of-Neural-Operators}.


\bibliography{sn-bibliography}
\end{document}